\documentclass[letterpaper, 10 pt, conference]{ieeeconf}  

\IEEEoverridecommandlockouts                              

\usepackage{graphics} 
\usepackage{epsfig} 
\usepackage{mathptmx} 
\usepackage{times} 
\usepackage{amsmath} 
\usepackage{amssymb}  

\usepackage{color}
\usepackage{xcolor}
\usepackage{caption}
\usepackage{preambles}

\title{\LARGE \bf
 Hybrid Zero Dynamics Inspired Feedback Control Policy Design for 3D Bipedal Locomotion using Reinforcement Learning}
\author{Guillermo A. Castillo$^{1}$, Bowen Weng$^{1}$, Wei Zhang$^{2}$, and Ayonga Hereid$^{3}$
\thanks{*This work was supported in part by the National Science Foundation under grant CNS-1552838 and the OSU M\&MS Discovery Theme Initiative.}
\thanks{$^{1}$Electrical and Computer Engineering, Ohio State University, Columbus, OH, USA;  {\tt\footnotesize \{castillomartinez.2, weng.172\}@osu.edu.}}
\thanks{$^{2}$SUSTech Institute of Robotics, Southern University of Science and Technology, China; {\tt\footnotesize zhangw3@sustech.edu.cn.}}
\thanks{$^{3}$Mechanical and Aerospace Engineering, Ohio State University, Columbus, OH, USA. {\tt\footnotesize hereid.1@osu.edu.}}%
}

\usepackage[capitalise]{cleveref}

\usepackage[noadjust]{cite}

\usepackage{graphicx}

\usepackage[caption=false]{subfig}

\begin{document}
\maketitle
\thispagestyle{empty}
\pagestyle{empty}

\begin{abstract}
This paper presents a novel model-free reinforcement learning (RL) framework to design feedback control policies for 3D bipedal walking. Existing RL algorithms are often trained in an end-to-end manner or rely on prior knowledge of some reference joint trajectories. Different from these studies, we propose a novel policy structure that appropriately incorporates physical insights gained from the hybrid nature of the walking dynamics and the well-established hybrid zero dynamics approach for 3D bipedal walking. As a result, the overall RL framework has several key advantages, including lightweight network structure, short training time, and less dependence on prior knowledge.  
We demonstrate the effectiveness of the proposed method on Cassie, a challenging 3D bipedal robot. The proposed solution produces stable limit walking cycles that can track various walking speed in different directions. Surprisingly, without specifically trained with disturbances to achieve robustness, it also performs robustly against various adversarial forces applied to the torso towards both the forward and the backward directions.
\end{abstract}

\section{Introduction}

3D bipedal walking is a challenging problem due to the multi-phase and hybrid nature of legged locomotion. Properties like underactuation, unilateral ground contacts, nonlinear dynamics, and high degrees of freedom significantly increase the model complexity. Existing approaches on bipedal walking can be roughly grouped into two categories: model-based and model-free methods. In~\cite{Grizzle2014Models}, the authors provide a comprehensive review of model-based methods, feedback control, and open problems of 3D bipedal walking. One of the main challenges for model-based methods is the limitation of mathematical models that capture the complex dynamics of a 3D robot in the real world. This results in non-robust controllers that require additional heuristic compensations and tuning processes, which can be time-consuming and requires experiences. 

Reduced order models, such as Linear Inverted Pendulum and its variants \cite{Kajita2001invertedpendulum}, have been studied extensively in the literature. For these simple models, stable walking conditions can be stated in terms of the ZMP (zero moment point) \cite{Vukobratovic2004ZMP, Yoshida2008, Stephens2010} or CP (capture point) \cite{Pratt2006, Pratt2012}, which can significantly simplify the control design. However, these approaches rely on some strong assumptions that often lead to quasi-static and unrealistic walking behaviors. 
Optimization-based methods such as Linear Quadratic Regulator (LQR)~\cite{Posa2016Optimization}, Model Predictive Control (MPC)~\cite{Erez2013Integrated, Koenemann2015Whole}, and Hybrid Zero Dynamics (HZD)~\cite{Westervelt2007Feedback, Ames2014Human} use the full order model of the robot to capture the underlying dynamics more accurately, which yields more natural dynamic walking behaviors. In particular, HZD is a formal framework for the control of bipedal robots with or without underactuation through the design of nonlinear feedback controllers and a set of virtual constraints. It has been successfully implemented in several physical robots, including many underactuated robots~\cite{Chevallereau2003RABBIT, Sreenath2011compliant, Zhao2017Multi, Hereid2018Dynamic}. Nonetheless, these methods are computationally expensive and sensitive to model parameters and environmental changes. Particularly for 3D walking, additional feedback regulation controllers are required to stabilize the system~\cite{Rezazadeh2015Spring, Reher2016Algorithmic, Gong2019Feedback}. Notably, recent work has successfully realized robust 3D bipedal locomotion by combining Supervised Learning with HZD \cite{Da2019Combining}.


\begin{figure}[tb]
  \centering
  \includegraphics[width=1\columnwidth, keepaspectratio]{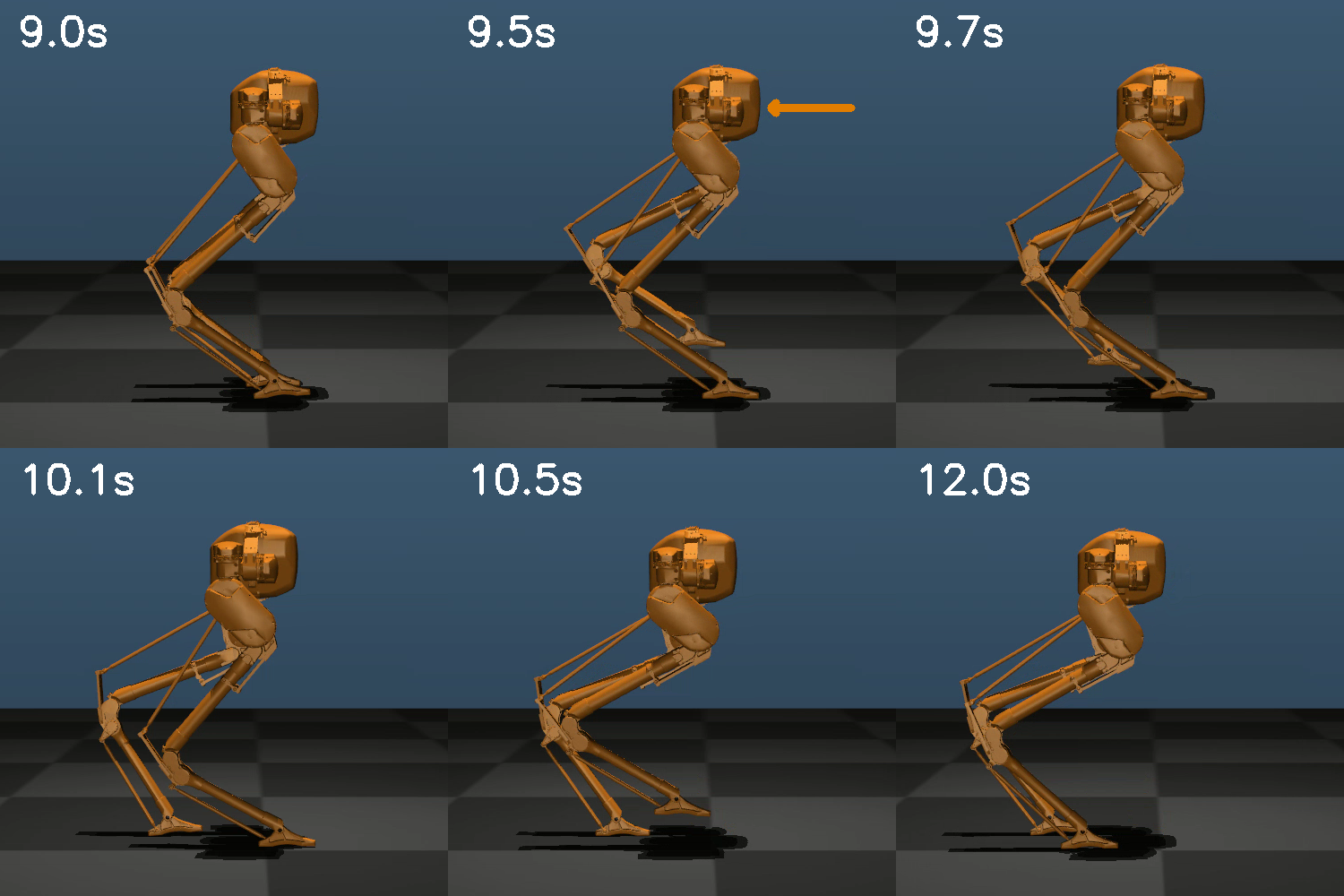}
  \caption{Cassie in simulation: gait recovering from the backward adversarial force with our proposed method.}
  \label{fig:cassie}
  \vspace{-5mm}
\end{figure}



With recent progress on deep learning, Reinforcement Learning (RL) has become a popular tool in solving challenging control problems in robotics. Existing RL methods often rely on end-to-end training without considering the underlying physics of the particular robot. A NN function is trained with policy gradient methods that directly maps the state space to a set of continuous actions~\cite{Lillicrap2015Continuous, schulman2017proximal}. Despite the empirical success, such methods are often sampling inefficient (millions of data samples) and are usually over-parameterized (thousands of tunable parameters). They may also lead to non-smooth control signals and unnatural motions that are not applicable to real robots. Through incorporating HZD with RL training,~\cite{Castillo2019Reinforcement} generated feasible trajectories that are tracked by PD controllers to produce sustainable walking gaits at different speeds. However, this method only works for a simple 2D robot model. For the more complex 3D bipedal walking,~\cite{xie2018feedback} adopts RL methods as part of the feedback control. The method relies on prior knowledge of a good joint reference trajectory and only learns small compensations added to the known reference trajectory, which does not provide the overall control solution. An imitation learning inspired method is proposed for a 3D robot~\cite{Xie2019Iterative}, which also requires a known walking policy and gradually improves the policy through learning. 

In this paper, we propose a novel hybrid control structure for robust and stable 3D bipedal locomotion. It harnesses the advantages of parameterized policies obtained through RL while exploiting the structure of the intuitive yet powerful additional regulations commonly used in 3D bipedal walking. The regulation terms embedded in the design of policy structure and reward functions differentiate our proposed method from the previous work of~\cite{Castillo2019Reinforcement} that is also inspired by HZD. We evaluate the performance of our method on Cassie, which is also a much more challenging robot than the 2D Rabbit~\cite{Castillo2019Reinforcement}. We further summarize the primary contributions of the present paper as follows:
\begin{itemize}
    \item \textbf{Model-free:} The trained policy naturally learns a feasible walking gait from scratch without the need of a given reference trajectory or the model dynamics. To the best of our knowledge, this is the first time that a variable speed controller for a 3D robot is learned without using previously known reference trajectories or training separate policies for different speeds.
    \item \textbf{Efficiency:} By incorporating the physical insight of bipedal walking, such as its hybrid nature, symmetric motion, and heuristic compensation, into the control structure and learning process, we significantly simplify the design to a shallow neural network (NN) with only 5069 trainable parameters. To the best of our knowledge, this is the smallest NN ever reported for Cassie. As a result, the NN policy is easy to train. It is also fast enough for real-time control with 1 kHz frequency on single process CPU (the low-level PD controller runs at 2 kHz).
    \item \textbf{Robustness:} The learned controller is capable of stably tracking a wide range of walking speeds in both longitudinal and lateral directions with just one trained policy. The robustness of the controller is also evaluated by several disturbance rejection tests in simulation. 
\end{itemize}


\section{PROBLEM FORMULATION} \label{section2}


In this section, we will review the classic HZD based feedback controllers for dynamic 3D walking robots. Inspired by the HZD framework, we then propose to study a model-free control problem using RL techniques. 

\subsection{Existing Challenges of the HZD Framework}
One of the main challenges of 3D bipedal walking is to find feasible trajectories that render stable and robust limit walking cycles while keeping certain desired behaviors, such as walking speeds of the system. 
In the HZD framework, to obtain such trajectories, an offline optimization problem is solved using the full-order model, and virtual constraints are introduced as a means to synthesize feedback controllers that realize stable and dynamic locomotion. By designing virtual constraints that are invariant through impact, an invariant sub-manifold is created---termed the \emph{hybrid zero dynamics surface}---wherein the evolution of the system is dictated by the reduced-dimensional dynamics of the under-actuated degrees of freedom of the system~\cite{Westervelt2007Feedback, Ames2014Human}. 

\begin{figure}
  \centering
  \vspace{1mm}\includegraphics[width=1\columnwidth, keepaspectratio]{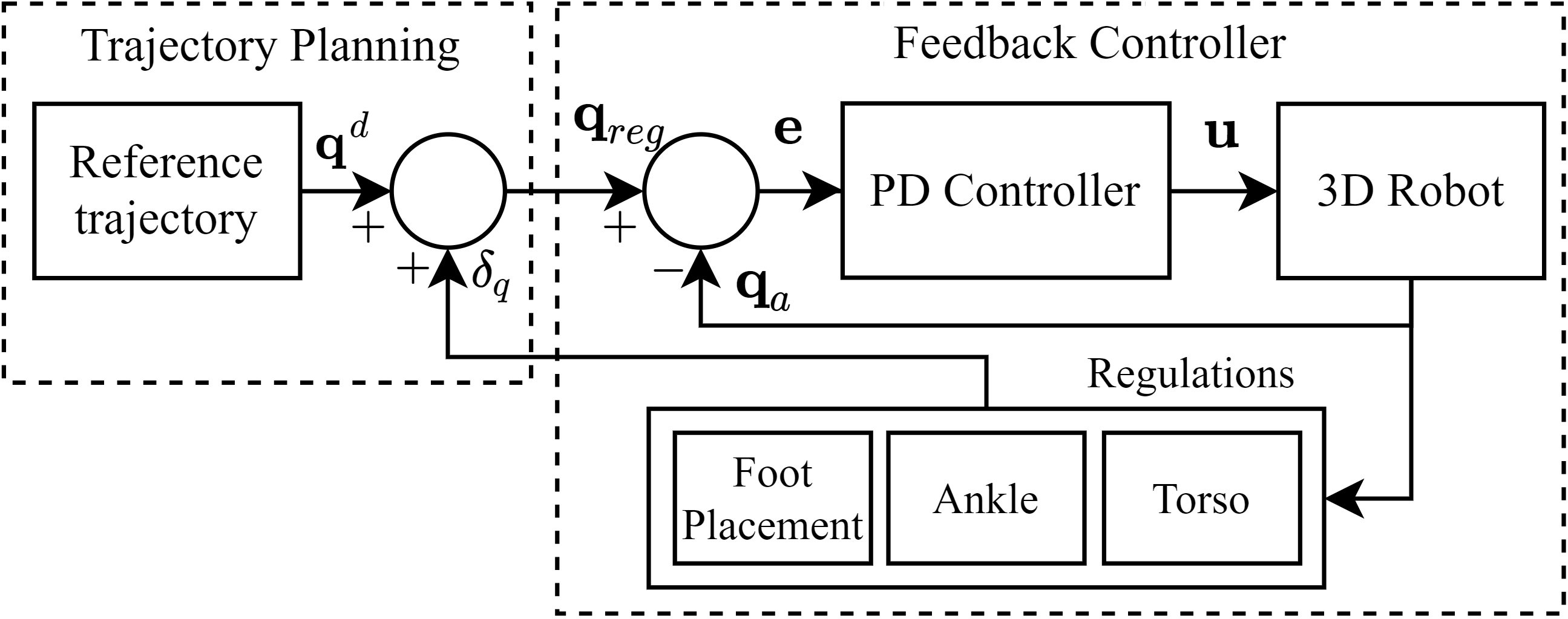}
  \caption{An example of 3D bipedal walking controller with heuristic feedback regulations.}
  \label{fig:control}
\vspace{-6mm}
\end{figure}

However, the mathematical models used in this optimization cannot completely capture the complex dynamics of a 3D bipedal robot. Consequently, additional heuristic compensation controllers or regulators are often required on top of the PD tracking controllers to stabilize the robot~\cite{Gong2019Feedback, Reher2016Algorithmic}. An example of such a control structure with feedback regulations for a 3D walking robot is shown in \figref{fig:control}. The compensations $\delta_{q}$ will either modify the original reference trajectories $\mathbf{q}^d$, or exert extra feed-forward torques based on additional feedback information, such as the lateral hip velocity or torso orientation, to improve the stability and robustness of the walking gaits in experiments. The new regulated reference trajectory $\mathbf{q}_{reg}$ is then tracked by PD controller through the control action $\mathbf{u}$.



\subsection{Structure of HZD-Based Feedback Controller} \label{section2feedback}
From a high-level abstraction, the problem of 3D bipedal walking can be divided in two stages: trajectory planning and feedback control.

\newsec{Virtual Constraints} Let $\mathbf{q}$ be the vector of joint coordinates of a general 3D bipedal robot, and  $\tau(t) \in [0,1] $ be a time-based phase variable (see \eqref{eq:tau} for explicit definition), then the virtual constraints are defined as the difference between the actual and desired outputs of the robot~\cite{Ames2013Human}:
\begin{align}
  \mathbf{y}_{2} &:= \mathbf{y}^a_{2}(\mathbf{q}) - \mathbf{y}^d_{2}(\tau(t),\alpha),
  \label{eq_vc}
\end{align}
where $\mathbf{y}^d_{2}$ is a vector of desired outputs defined in terms of $5^{\text{th}}$ order B\'ezier polynomials parameterized by the coefficients $\mathbf{\alpha}$, given as:
\begin{align}
    \mathbf{y}^d_2(\tau(t),\mathbf{\alpha}) := \sum_{k=0}^{5} \alpha[k] \frac{M!}{k!(M-k)!} \tau(t)^k (1-\tau(t))^{M-k}.
\end{align}

In this paper, we choose $\tau(t)$ to be the scaled relative time with respect to step time interval, i.e.,
\begin{align}
    \label{eq:tau}
    \tau(t) = \frac{t - t^-}{t_{step}},
\end{align}
where $t_{step}$ is the duration of one walking step, and $t^-$  is the time at the beggining of the step. It is important to denote that by properly choosing the coefficients of these B\'ezier polynomials, one can achieve different walking motions.

In the HZD framework, the B\'ezier coefficients are obtained from the solution of an optimization problem whose cost function and constraints are determined by the desired behavior of the robot. Then, the B\'ezier Polynomials define the desired trajectories to be tracked in order to drive the virtual constraints to zero. However, for 3D bipedal walking robots, simply tracking desired trajectories is not enough to achieve a stable walking motion. Therefore, the following regulations are added to the controller: foot placement, torso regulation, and ankle regulation.

\newsec{Foot placement} controller has been widely used in 3D bipedal walking robots with the objective of improving the speed tracking and the stability and robustness of the walking gait~\cite{Da2016, Rezazadeh2015Spring, Gong2019Feedback}. Longitudinal speed regulation, defined by \eqref{eq:long_reg}, sets a target offset in the swing hip pitch joint, whereas lateral speed regulation \eqref{eq:lat_reg} do the same for the swing hip roll angle. In these equations, $v_{x}[k]$ and $v_{y}[k]$ are the average longitudinal and lateral speeds of the robot at the middle of step $k$, $v^{d}_{x}$, $v^{d}_{x}$ are the reference speeds, and $K_{p_{x}}, K_{d_{x}}, K_{p_{y}}, K_{d_{y}}$ are manually tuned gains. 
\begin{align}
    \label{eq:long_reg}
    \hspace{-3em}
    \delta^{sw}_{hpitch}[k] &= K_{p_{x}}(v_{x}[k]-v^{d}_{x}) + K_{d_{x}}(v_{x}[k]-v_{x}[k-1]),\\
    \label{eq:lat_reg}
    \delta^{sw}_{hroll}[k] &= K_{p_{y}}(v_{y}[k]-v^{d}_{y}) + K_{d_{y}}(v_{y}[k]-v_{y}[k-1]).
\end{align}

\newsec{Torso regulation} is applied to keep the torso in an upright position, which is desired for a stable walking gait. Assuming that the robot has a rigid body torso, simple PD controllers defined by \eqref{eq:torso_roll} and \eqref{eq:torso_pitch} can be applied respectively to the hip roll and hip pitch angle of the stance leg:
 \begin{align}
    \label{eq:torso_roll}
    u^{st}_{hroll} &= K_{p_{troll}}(\phi - \phi^{d}) + K_{d_{troll}}(\dot{\phi} - \dot{\phi}^{d}),\\
    \label{eq:torso_pitch}
    u^{st}_{hpitch} &= K_{p_{tpitch}}(\theta - \theta^{d}) + K_{d_{tpitch}}(\dot{\theta} - \dot{\theta}^{d}),
\end{align}
where $\phi$ and $\theta$ are the torso roll and pitch angles, and $K_{p_{troll}}, K_{d_{troll}}, K_{p_{tpitch}}, K_{d_{tpitch}}$ are manually tuned gains. 
 
\newsec{Ankle regulation} is applied to keep the swing foot flat during the whole swinging phase, including the landing moment. For Cassie, this can be done by using forward kinematics for the reference trajectory of the swing ankle joint, given as
  \begin{align}
    \label{eq:flat_foot_reg}
    \gamma^{sw} = \theta - 13\deg - 50\deg, 
\end{align}
where $\gamma^{sw}$ is the ankle joint corresponding to the pitch angle of the swing foot. 
In addition, to stabilize the walking gait, especially when walking on soft surfaces~\cite{Gong2019Feedback}, the stance foot pitch angle of the robot will be set to be passive. 

It is important to denote that the speed and torso regulations presented above are fixed, intuitive, and applicable to any general 3D bipedal walking robot. However, given the decoupled structure of the controller used for the different regulations, there are several gains that need to be manually tuned in order to achieve improved stability, which is time-consuming and requires experience. However, this process can be easily automated within an RL framework.

\subsection{Tackling the Problem with Reinforcement Learning}



Inspired by the nice properties of HZD (low-dimensional space, accounting of the hybrid nature of walking, virtual constraints), we propose an RL framework that incorporates those properties in the learning process to solve the complex problem of 3D walking. The main objective is to create a unified policy that can handle both problems presented in Fig. \ref{fig:control}: trajectory planning and feedback control.

By this, we aim to address two specific challenges generally present in the current methods for 3D bipedal walking (i) eliminating the hideous task of manually tuning the gains of the feedback regulations by including them into the learning process, and (ii) improving the data efficiency of the RL method by reducing significantly its number of parameters.

To validate the proposed approach, we use Cassie-series bipedal robot, designed by Agility Robotics, as our test-bed in this paper. This underactuated biped has 20 degrees of freedom (DOF) in total. Each leg has seven joints, in which five of them are directly actuated by electrical motors and the other two joints are connected via specially designed leaf-spring four-bar linkages for additional compliance. 
When supported on one foot during walking, the robot is underactuated due to its narrow feet. Agility Robotics has released dynamic simulation models of the robot in MuJuCo~\cite{AgilitySimsJune2018}, which will be used later in this paper.

\section{APPROACH} \label{section3}
In this section, we propose a non-conventional RL framework that combines a learning structure inspired by the HZD with the foot placement, torso and ankle regulation introduced in section \ref{section2}. We incorporate useful insights from traditional control framework into the learning process of the control policy. By HZD-inspired, we refer to the fact that the learning structure uses a low dimensional representation of the state and a time-based phase variable to command the behavior of the whole system while enforcing invariance of the virtual constraints through impact and symmetry conditions of the walking gait. 


\subsection{Overall Framework}
We formally present a non-conventional RL framework that uses a low dimensional state of the robot to learn a robust control policy able to track different walking speeds while maintaining the stability of the walking limit cycle. The proposed framework first establishes a NN function that maps a reduced order of the robot's state to (i) a set of coefficients of the B\'ezier polynomials that define the trajectory of the actuated joints, and (ii) a set of gains corresponding to the derivative gain of the joints PD controller, as well as the gains for the foot placement and torso regulations described in section \ref{section2feedback}. Independent low level PD controllers are then used to track the desired output for each joint, which enforces the compliance of the HZD virtual constraints.

\begin{figure}
\centering
\includegraphics[width=8.5cm]{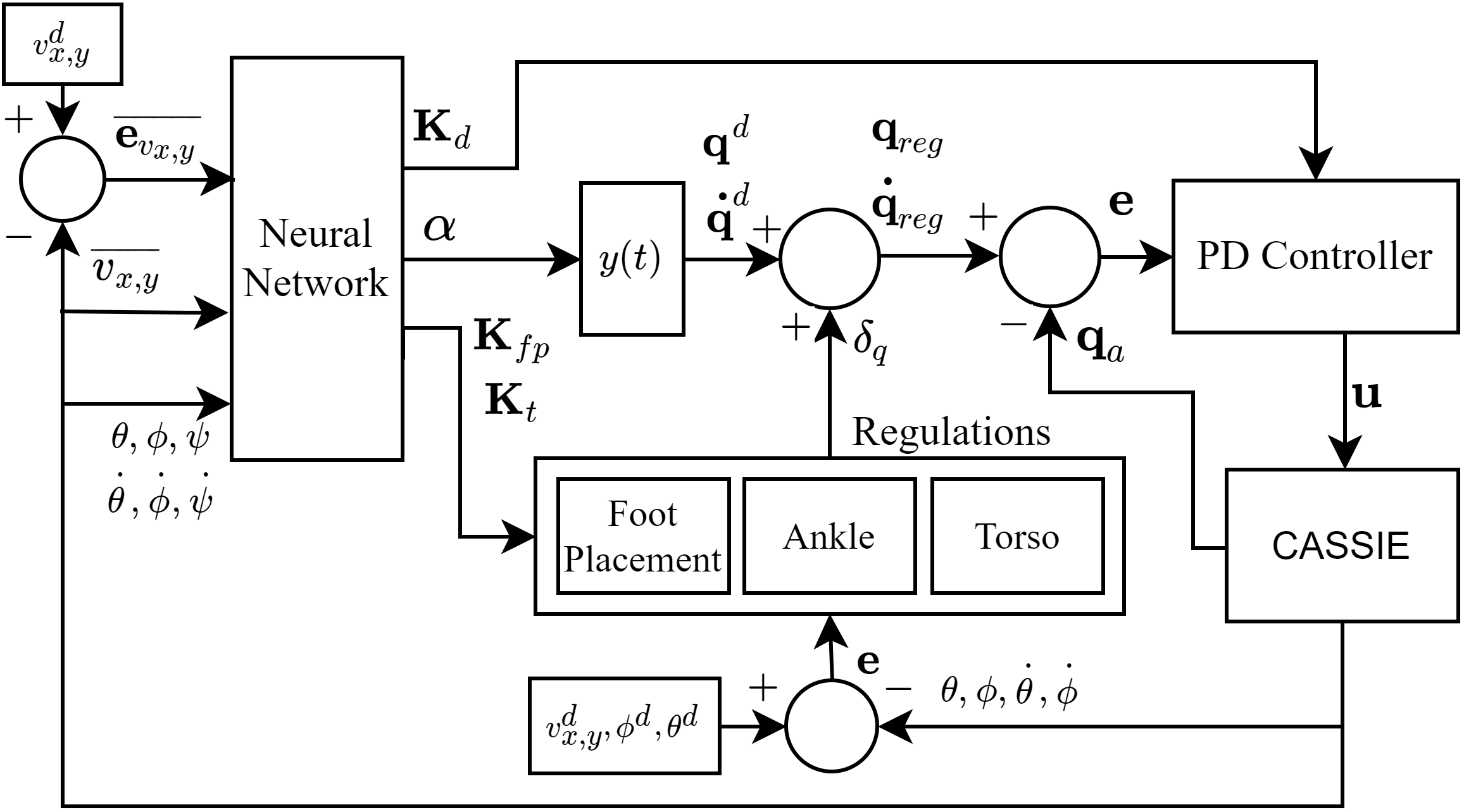}
\caption{Overall structure of the RL framework.}
\label{fig:structure}
\vspace{-4mm}
\end{figure}

A diagram of the overall RL framework is presented in \figref{fig:structure}. At each time step, the trained policy maps the inputs of the desired walking velocity, the average actual velocity, the average velocity tracking error, and the torso orientation and angular velocity to the set of coefficients $\alpha$, the derivative gains of the joint PD controller, and the compensation gains for the foot placement and torso regulation. A detailed explanation of the NN structure will be given in Section \ref{subsec:NN_structure}. Then, $\alpha$ is used jointly with the phase variable $\tau(t)$ to compute each joint's desired position and velocity. Compensation gains are used in the foot placement regulation and torso regulation to compute the trajectory compensation for hip roll and pitch angle of the stance and swing leg of the robot. In addition, the ankle regulation computes the trajectory for the swing and stance leg ankle joints. The PD controller uses the tracking error between the desired and actual value of the output to compute the torque of each actuated joint, which is the input of the dynamic system that represents the walking motion of the robot. To close the control loop, the measurement of the robot's states are used as feedback for the inner and outer control loops.

It is worth mentioning that the reference trajectories are learned from scratch and naturally obtained by the proposed RL framework. This is in contrast to some existing studies of RL~\cite{Peng2017DeepLoco}, \cite{xie2018feedback},~\cite{Xie2019Iterative}, which rely on some given working policy providing the joints reference trajectories. 


\subsection{Neural Network Structure} \label{subsec:NN_structure}
\begin{figure}[t]
\centering
\includegraphics[width=8cm]{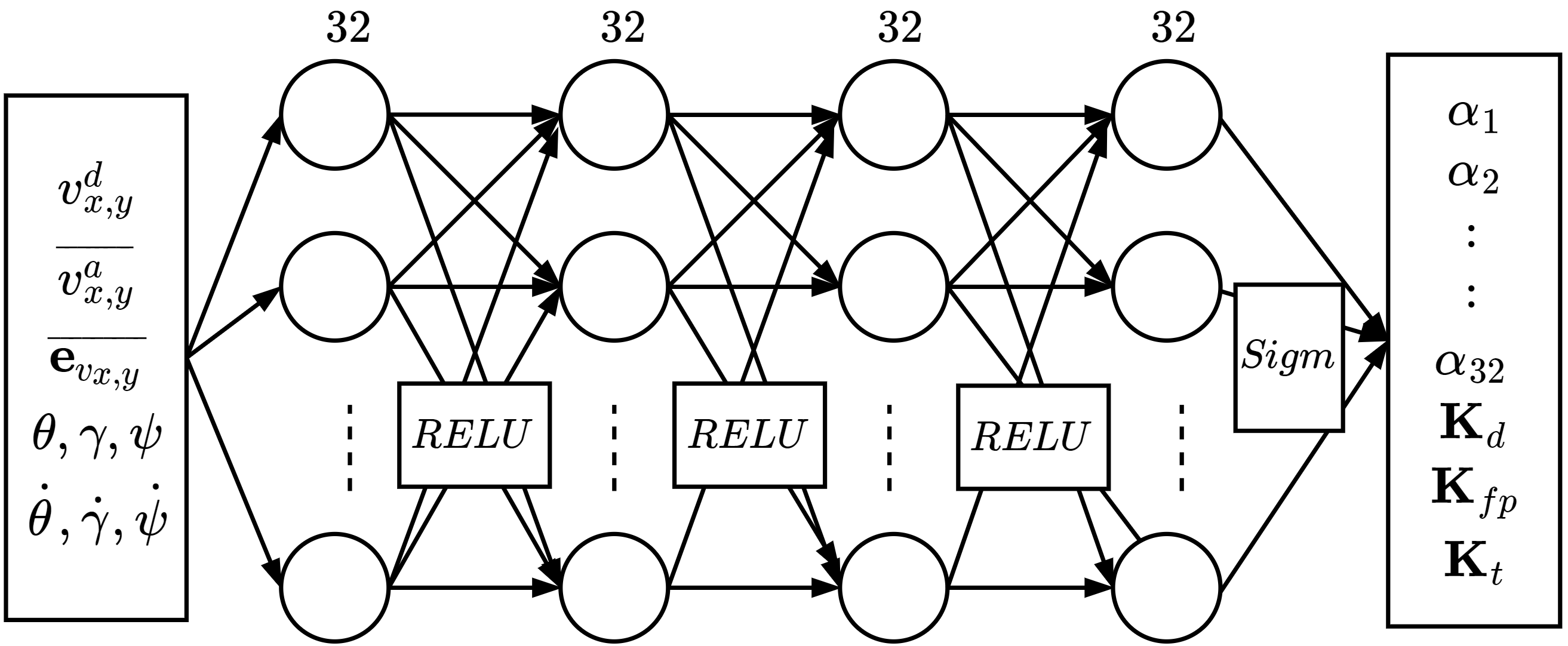}
\caption{Structure of the neural network.}
\label{fig:NN}
\vspace*{-4mm}
\end{figure}

\figref{fig:NN} shows the structure of the NN implemented for the learning process. Cassie's dynamics model contains 40 states, the robot's pelvis position, velocity, orientation, and angular velocity, plus the angle and angular velocity of all the active and passive joints of the robot. However, the proposed NN only contains 12 dimensional reduced-order state: desired longitudinal and lateral velocity ($v^d_x$, $v^d_y$), average longitudinal and lateral velocity ($\overline{v_x}$ $\overline{v_y}$), average longitudinal and lateral velocity error ($\overline{e_{v_x}}$, $\overline{e_{v_y}}$), roll, pitch and yaw angles ($\phi,\theta,\psi$), and roll, pitch and yaw angular velocities ($\dot{\phi},\dot{\theta},\dot{\psi}$). Here, we consider the average speed as the speed during one walking step of the robot, which takes about 350 $ms$.  All the inputs are normalized in the interval $[-0.5, 0.5]$. The value of the desired velocity is uniformly sampled from a continuous space interval from -0.5 to 1.0~$m/s$.

The output of the NN corresponds to the coefficients of the B\'ezier polynomials, denoted by $\alpha$, and the set of gains of the PD controller, foot placement and torso compensations, denoted by $\mathbf{K}_{d}$, $\mathbf{K}_{fp}$ and $\mathbf{K}_t$, respectively. Initially, since the robot has ten actuated joints and each B\'ezier polynomial is of degree 5, the total size of the set of parameters $\alpha$ should be 60 for the right stance and 60 for the left stance. In addition, the same derivative gains $\mathbf{K}_d$ are used for the corresponding joints of the left and right legs, which is possible because of the symmetric nature of the walking gait. Finally, by equations \eqref{eq:long_reg}-\eqref{eq:torso_roll}, $\mathbf{K}_{fp}$ and $\mathbf{K}_t$ are of the form
\begin{equation}
    \begin{aligned}
        \label{eq:K_fp}
        \mathbf{K}_{fp} &= [K_{p_{x}}, K_{d_{x}}, K_{p_{y}}, K_{d_{y}}], \\
        \mathbf{K}_{t} &= [K_{p_{troll}}, K_{d_{troll}}, K_{p_{tpitch}}, K_{d_{tpitch}}].
    \end{aligned}
\end{equation}    

Then $\mathbf{K}_{d}$ is of dimension 5, and both, $\mathbf{K}_{fp}$ and $\mathbf{K}_t$, are of dimension 4. This results in a total of 133 outputs. Nonetheless, by considering the physical insight of the dynamic walking, we can significantly reduce the number of outputs of the NN from 128 to 45. This will be explained in detail in Section \ref{subsec:reduction}. The number of hidden layers of the NN is 4, each with 32 neurons. ReLU activation functions are used between hidden layers, whereas the final layer employs a sigmoid function to limit the range of the outputs. As compared with other methods in the literature~\cite{Lillicrap2015Continuous,xie2018feedback}, the proposed NN is much smaller in size, making the overall RL method sample efficient and easy to implement. 

Finally, due to the properties of the family of polynomials used to parameterize the joints trajectories, the set of B\'ezier coefficients accurately define the upper and lower bounds of the desired output trajectories. That is, for each set of B\'ezier coefficients $\alpha_i$ and desired trajectory $q^d_i$ associated with the $i^{th}$ joint, we have 
\begin{align}
    \label{eq:traj_bounds}
    q^{min}_i < \alpha_i^{min} < q^{d}_i < \alpha_i^{max} < q^{max}_i.
\end{align}
Therefore, the output range of the set of parameters can be limited by the physical constraint of each actuated joints, or even more, by the expected behavior of the robot ($q^{min}_i, q^{max}_i $). This critical feature significantly reduces the continuous interval of the output, which decreases the complexity of the RL problem and improves the efficiency of the learning process. 

\subsection{Reduction of Output Dimension} \label{subsec:reduction}
For a general walking pattern, there exists a symmetry between the right and left stance. Therefore, given the set of coefficients for the right stance $\mathbf{\alpha}_R \in \mathbb{R}^{6x10}$, where each column represents the B\'ezier coefficients for a desired joint trajectory, we can easily obtain the set of coefficients for the left stance $\mathbf{\alpha}_L \in \mathbb{R}^{6x10}$ by 
\begin{align}
    \label{eq:symm_cond}
    \mathbf{\alpha}_L = \mathbf{\alpha}_R \mathbf{T}
\end{align}
where $\mathbf{T} \in \mathbb{R}^{10x10}$ is a very sparse transformation matrix that represents the symmetry between the joints of the right and left legs of the robot.

To encourage the smoothness of the control actions after the ground impact, we enforce that at the beginning of every step the initial point of the B\'ezier polynomial coincides with the current position of the robot's joints. That is, for each joint $i$ with Bézier coefficients $\mathbf{\alpha}_R \in \mathbb{R}^{6x10}$, we have
\begin{align}
    \label{eq:impact_cond}
    \alpha_i[0] = q_i(\tau(0)).
\end{align}
In addition, to encourage the invariance of the virtual constraints through impact, we enforce the position of the hip joints and knee joints to be equal at the beginning and end of the step ($\tau(t)=0$ and $\tau(t)=1$ respectively).

Finally, the ankle regulation enforces the stance ankle to be passive and the trajectory of the non-stance ankle to be defined by forward kinematics accordingly to \eqref{eq:flat_foot_reg}. Therefore, we do not need to find the B\'ezier coefficients for the stance and swing ankle joints.

\subsection{Learning Procedure}
Provided the NN policy structure and the reduced desired output of actions, the network is then trained with the evolution strategies (ES)~\cite{salimans2017evolution}. Note that our proposed method is not limited to a particular training method. It can be trained using any RL algorithms that can handle continuous action space, including evolution strategies (ES), proximal policy optimization (PPO)~\cite{schulman2017proximal}, and deterministic policy gradient methods~\cite{silver2014deterministic}.

In this paper, we adopt the following reward function in training for Cassie:
\begin{equation}
    \label{eq:reward func}
    r = \mathbf{w}^T \mathbf{r},
\end{equation}
with a vector of 8 customized rewards $\mathbf{r}$ and the weights $\mathbf{w}$. Specifically, 
\begin{equation}
    \mathbf{r} = [ r_{v_x}, r_{v_y}, r_h, r_{u}, r_{COM}, r_{ang}, r_{angvel}, r_{fd} ]^T.
\end{equation}
This encourages better velocity tracking (through $r_{v_x}, r_{v_y}$), height maintenance ($r_h$), energy efficiency ($r_{u}$) and natural walking gaits ($r_{COM}, r_{ang}, r_{angvel}, r_{fd}$). 
Starting from a random initial state that is close to a "stand-up" position with zero velocity and a uniformly sampled desired velocity, we collect a trajectory of states, actions and rewards, referred as an episode. 
The episode length is 10000 simulation steps and it has an early termination if any of the following conditions is violated: 
\begin{equation}
    \begin{aligned}
    &\psi|<0.5, \quad |\theta|<0.5, \quad |\phi|<0.5, \\ &\quad 0.75<p_z<1.1, \quad \Delta_f < 0.05,
    \end{aligned}
\end{equation}
where $p_z$ is the height of the robot's pelvis and $\Delta_f$ is the distance between the feet.


\section{SIMULATION RESULTS}
To validate the proposed method, a customized environment for Cassie was built in MuJoCo physics engine~\cite{Todorov2012MuJoCo}. We used the model information of Cassie robot provided by Agility Robotics and the Oregon State University, which is publicly available~\cite{AgilitySimsJune2018}. The number of trainable parameters for the NN is 5069, and the training time is about 10 hours using a single 12-core CPU machine. Visualized results of the learning process and evaluation of the policy in simulation can be seen in the accompanying video submission~\cite{video_link}. This section presents the performance of the control policy obtained from the training in terms of (i) speed tracking, (ii) disturbance rejection, and (iii) the convergence of stable periodic limit cycles. 


\subsection{Speed Tracking}
\begin{figure}
\centering
\vspace{1mm}
\includegraphics[trim={0cm 0cm 5cm 5cm},clip,width=1\columnwidth]{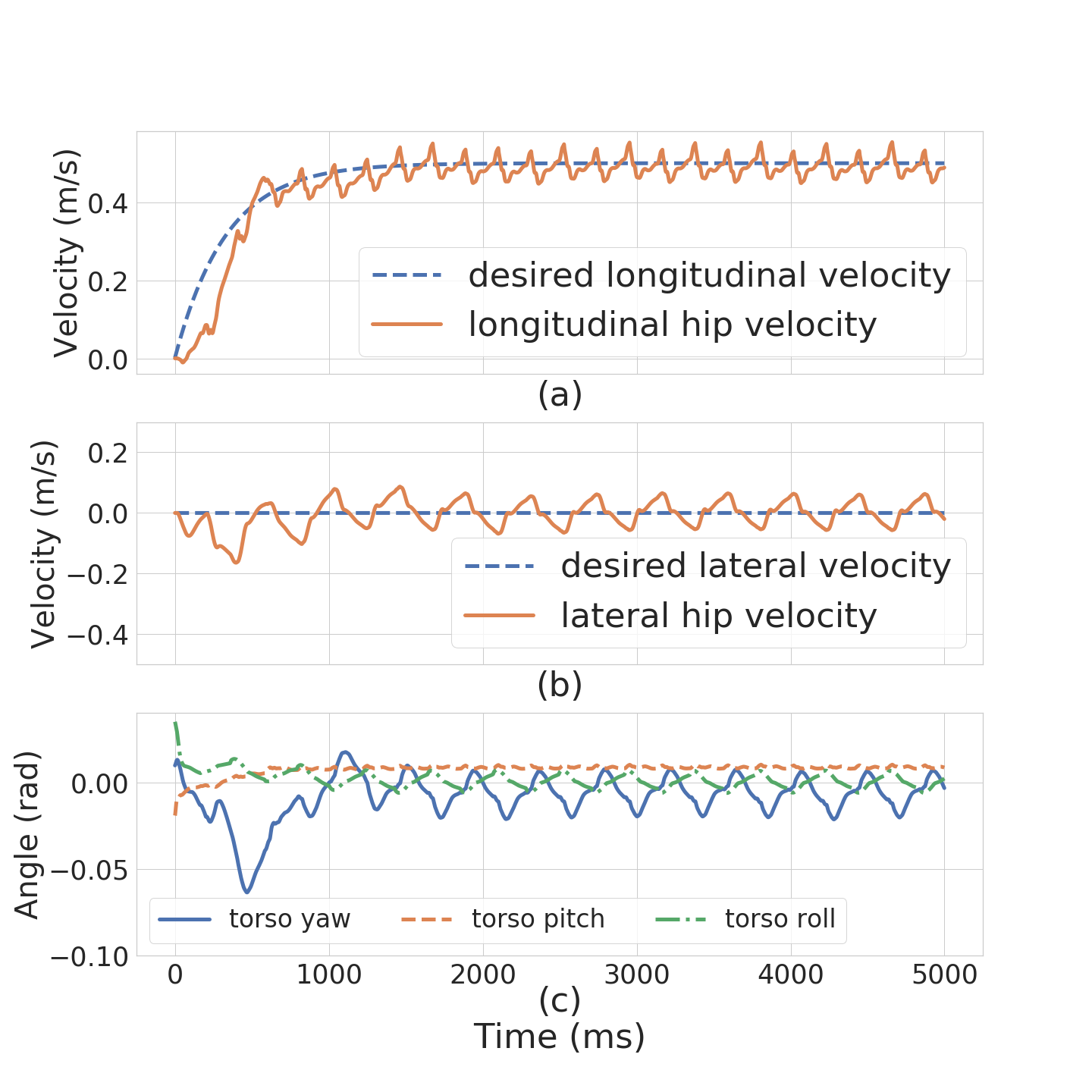}
\vspace{-7mm}
\caption{Performance of the learned policy while tracking a fixed desired longitudinal walking speed.} 
\label{fix_vel}
\vspace{-3mm}
\end{figure}

Due to the decoupled structure, the learned controller can effectively track a wide range of desired walking speeds in both longitudinal and lateral directions. The performance of tracking a fixed desired speed of 0.5 $m/s$ in the forward direction is shown in \figref{fix_vel}. From \figref{fix_vel}(c), one can see the controller is capable of keeping the upright position of the torso while walking. This particular behavior is encouraged by the reward function during the training process, and it also contributes to the stability of the walking gait.

\begin{figure}
\centering
\vspace{1mm}
\includegraphics[trim={0cm 1cm 5cm 4cm},clip,width=1\columnwidth]{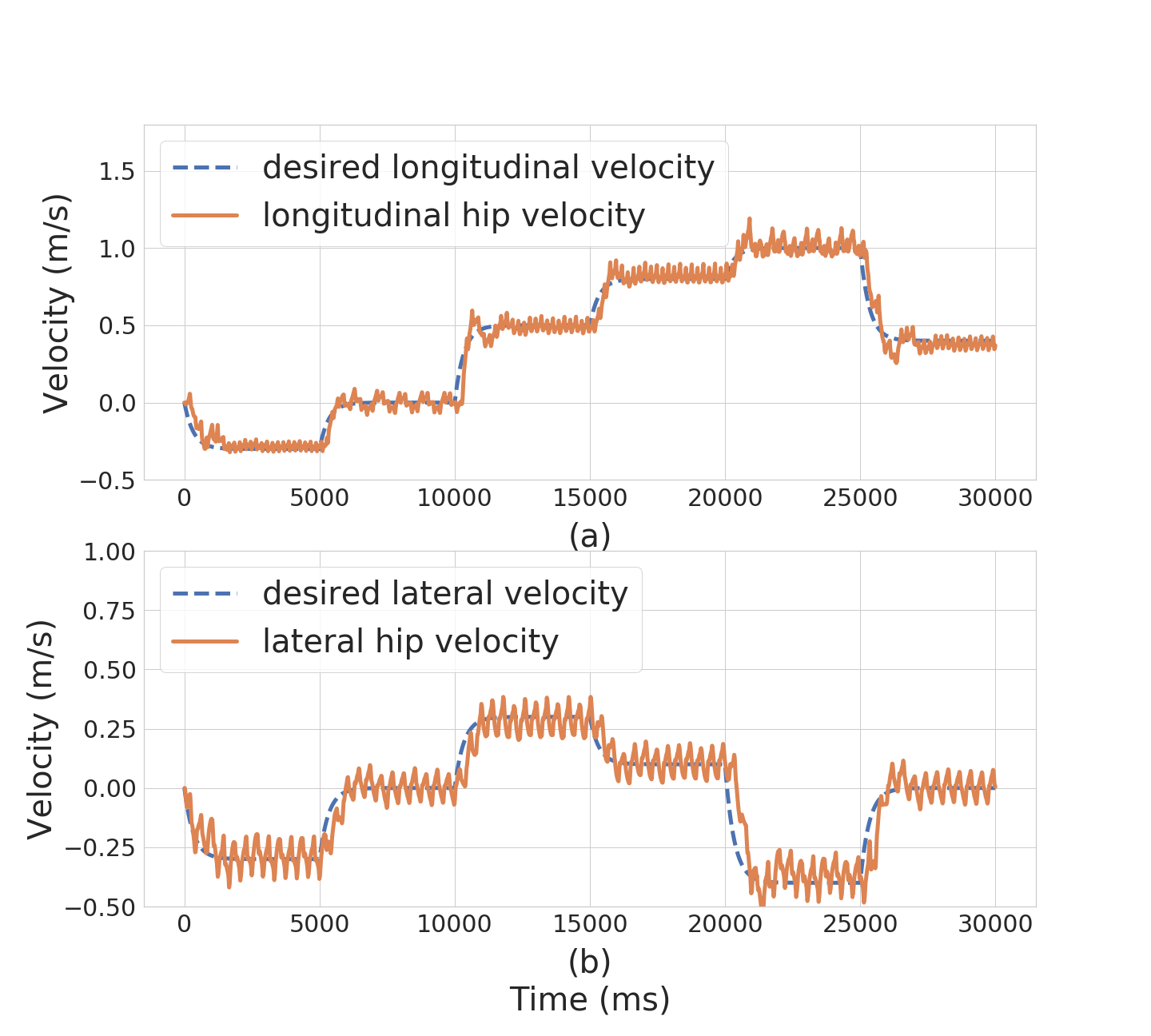}
\caption{Performance of the learned policy while tracking varying desired longitudinal and lateral walking speeds.}
\label{change_vel}
\vspace{-3mm}
\end{figure}

The performance of continuously tracking various desired speeds is shown in \figref{change_vel},  in an interval from -0.5 to 1.0 $m/s$ longitudinally ($v_x$), and an interval from -0.3 to 0.3 $m/s$ in the lateral direction ($v_y$). Note that walking at negative $v_x$ means the robot is walking backward, whereas moving at positive $v_y$ implies the robot is taking side steps to the left. In both cases, the controller is able to handle any speed change without falling or losing track of the reference, even after steep changes in the desired velocity.

\subsection{Disturbance Rejection}
\begin{figure}
\centering
\includegraphics[trim={4cm 1cm 5cm 3cm},clip,width=1\columnwidth]{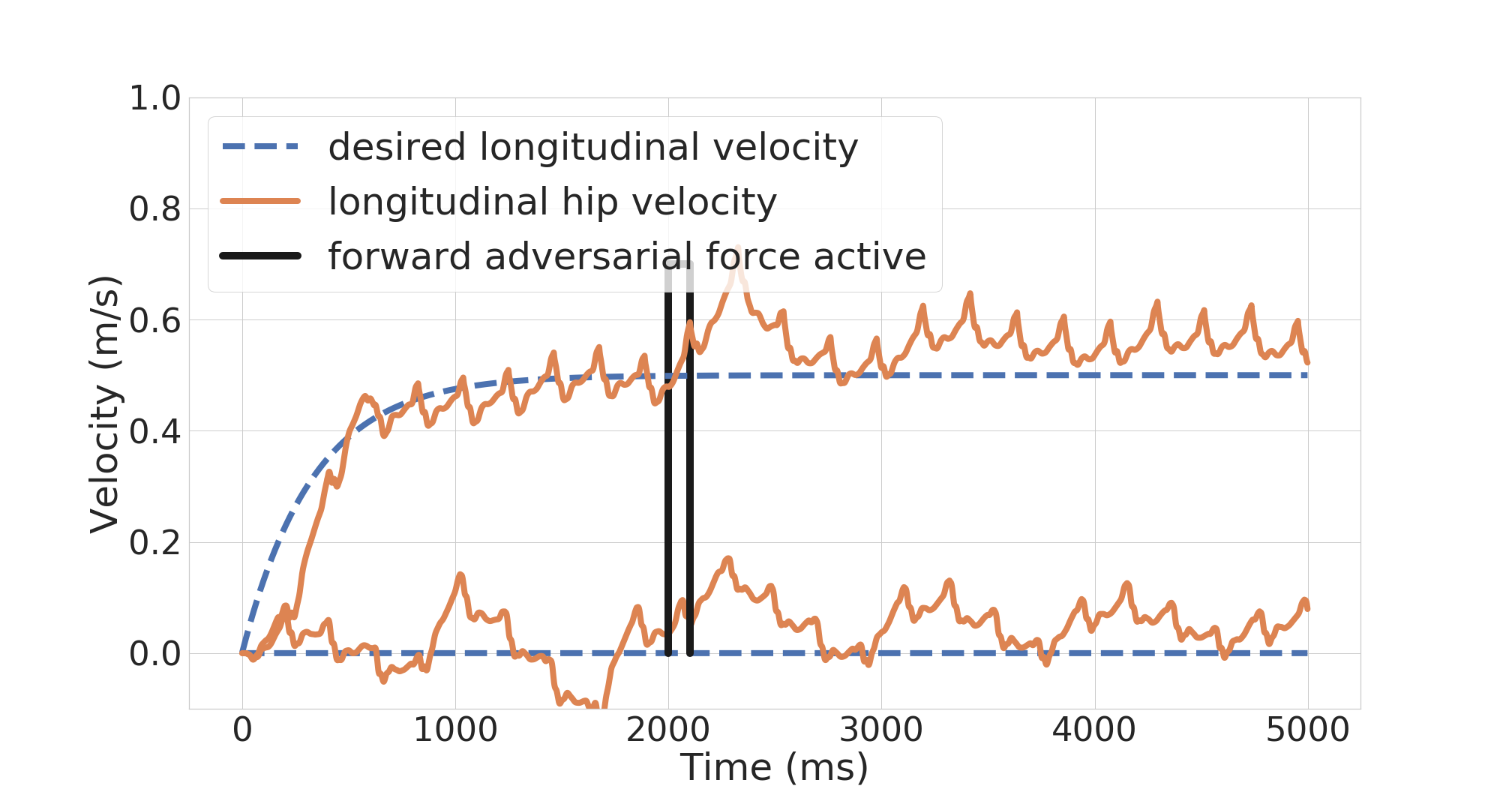}
\caption{Robustness of the controller when an adversarial force is applied in the forward direction.}
\label{fig:forward_adv}
\vspace{-3mm}
\end{figure}

\begin{figure}
\centering
\includegraphics[trim={1cm 1cm 5cm 3cm},clip,width=1\columnwidth]{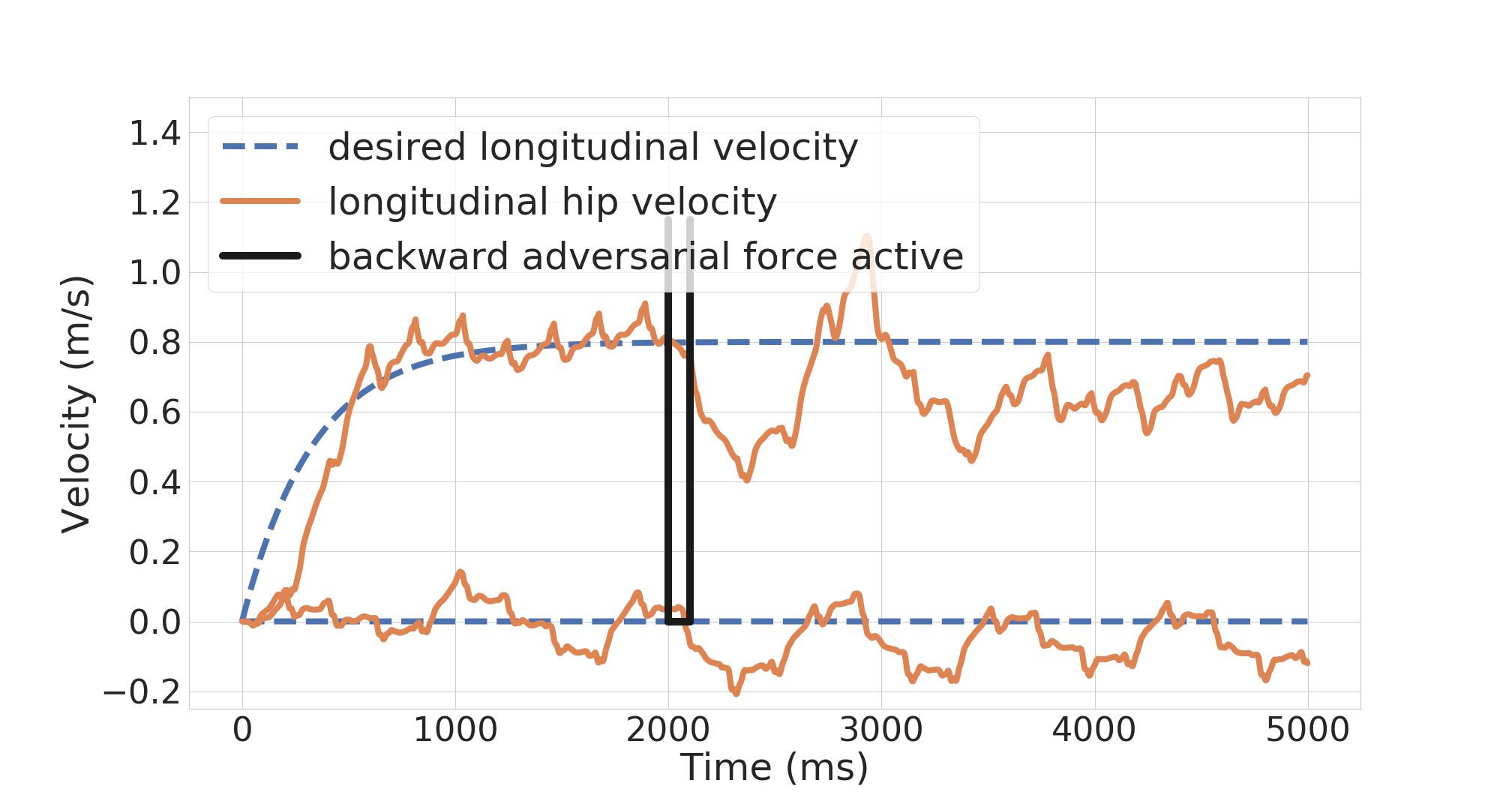}
\caption{Robustness of the controller when an adversarial force is applied in the backward direction.}
\label{fig:backward_adv}
\vspace{-3mm}
\end{figure}

To evaluate the robustness of our controller, we applied an adversarial force directly at the robot's pelvis in both the forward and the backward directions. It is worth emphasizing that we do not inject any torso disturbance throughout the training process. The robustness of the policy is achieved naturally through the constant updates of the B\'ezier coefficients, the derivative gains of the adaptive PD controller, and the gains of the additional foot placement, torso and ankle regulators. In the results shown in \figref{fig:forward_adv} and \figref{fig:backward_adv}, we adopt the adversarial force with the same magnitude of 25 $N$ in both directions. It is applied 2 seconds after starting the test and lasts for 0.1 seconds. Throughout our adversarial tests, the robot can handle up to 40 $N$ in the forward direction and 45 $N$ in the backward direction without falling, but the speed tracking may take a long time to recover with an external force of high magnitude.

\figref{fig:forward_adv} illustrates the response of the controller for a forward adversarial force when the robot is walking in place and walking forward at 0.5 $m/s$. \figref{fig:backward_adv} illustrates the response of the controller when the same force of 25 $N$ is applied in the backward direction while the robot is walking forward at 0 and 0.8 $m/s$. Throughout the four tests, the robot never falls and always closely recovers to the desired velocity.


\subsection{Periodic Stability of the Walking Gaits}
\begin{figure}
\centering
\includegraphics[trim={4cm 3cm 5cm 6cm},clip,width=1\columnwidth]{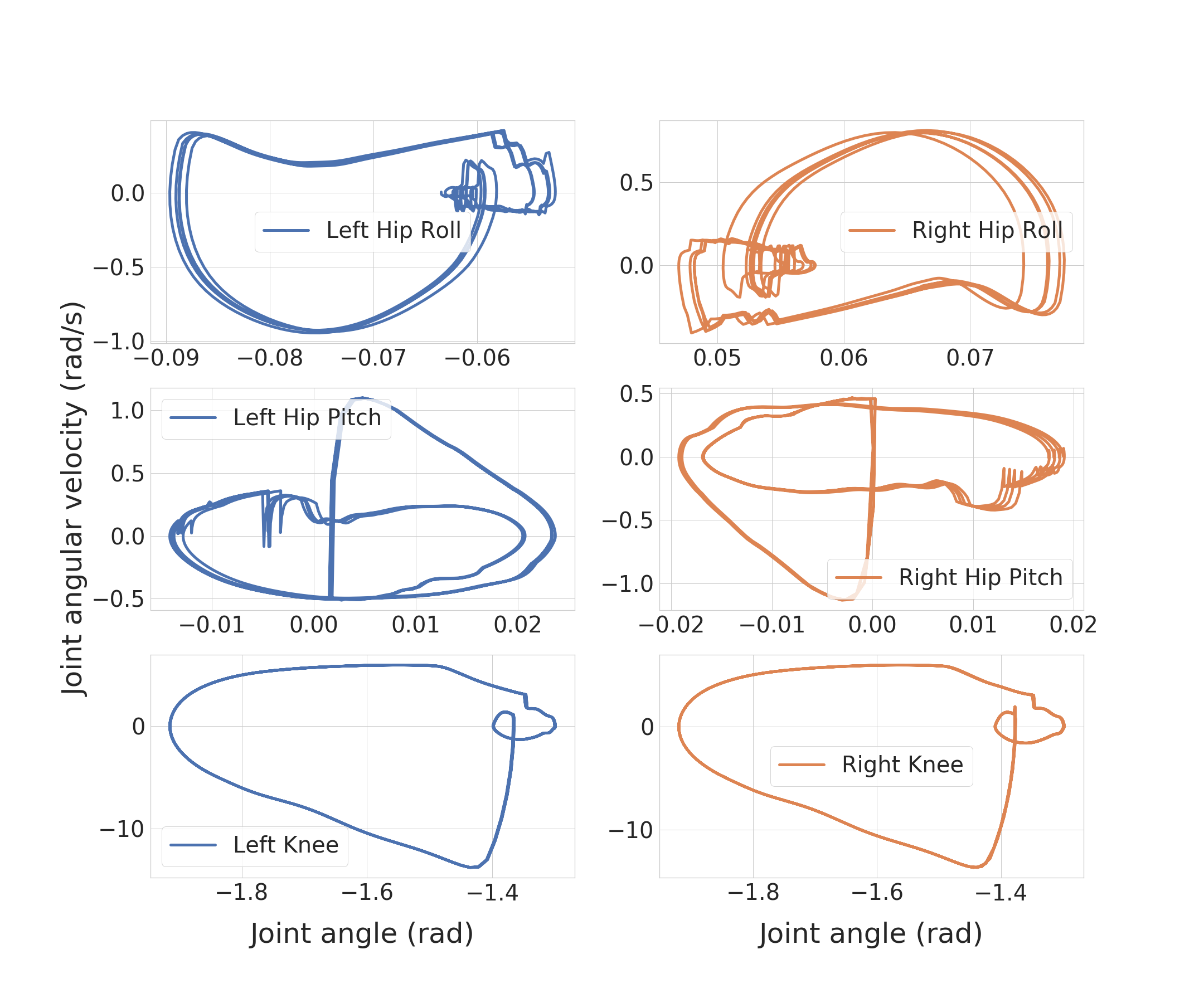}
\caption{Walking limit cycle of the learned policy with the desired longitudinal velocity of 0.5 $m/s$.}
\label{limit_cycle}
\vspace{-3mm}
\end{figure}

Periodic stability is one of the most important metrics for assessing the stability of walking gaits. In this paper, we only empirically evaluated the stability by observing the joint limit cycles of a periodic walking gait. \figref{limit_cycle} shows that the convergence of several representative robot actuated joints to periodic limit cycles during a fixed speed walking. Moreover, the orbit described by the left and right joints demonstrates the symmetry of walking gaits. This is due to the specific feature we encouraged in the design of the control policy.

\section{Conclusion}
This paper presents a novel model-free RL approach for the design of feedback controllers for 3D bipedal robots. The unique decoupled structure of the learned control policy incorporates the physical insights of the dynamic walking and heuristic compensations from classic 3D walking controllers. The result is a data-efficient RL method with a reduced number of parameters in the NN that can learn stable and robust dynamic walking gaits from scratch, without any reference motion or expert guidance. The learned policy demonstrates good velocity tracking and disturbance rejection performances on a 3D bipedal robot. 


\newpage
\bibliography{ms.bib}
\bibliographystyle{IEEEtran}

\end{document}